\documentclass[10pt,twocolumn,letterpaper]{article}

\usepackage{iccv}
\usepackage{times}
\usepackage{epsfig}
\usepackage{graphicx}
\usepackage{amsmath}
\usepackage{amssymb}
\usepackage{color}
\usepackage{tabulary,multirow,overpic,xcolor}
\usepackage{custom}

\definecolor{citecolor}{RGB}{34,139,34}
\definecolor{lightred}{RGB}{241,140,142}
\definecolor{citecolor2}{HTML}{0071bc}

\usepackage[pagebackref=true,breaklinks=true,letterpaper=true,colorlinks,citecolor=citecolor2,bookmarks=false]{hyperref}

\iccvfinalcopy 

\def\iccvPaperID{3096} 

\begin{document}

\title{Visual Alignment Constraint for Continuous Sign Language Recognition}

\author{Yuecong Min\textsuperscript{1,2}, 
Aiming Hao\textsuperscript{1,2}, 
Xiujuan Chai\textsuperscript{3}, 
Xilin Chen\textsuperscript{1,2}\\
\textsuperscript{1}Key Lab of Intelligent Information Processing of Chinese Academy of Sciences (CAS),\\
Institute of Computing Technology, CAS, Beijing, 100190, China\\
\textsuperscript{2}University of Chinese Academy of Sciences, Beijing, 100049, China\\
\textsuperscript{3}Agricultural Information Institute, Chinese Academy of Agricultural Sciences, Beijing, 100081, China\\
{\tt\small \{yuecong.min,aiming.hao\}@vipl.ict.ac.cn, chaixiujuan@caas.cn, xlchen@ict.ac.cn}}


\maketitle

\begin{abstract}
Vision-based Continuous Sign Language Recognition (CSLR) aims to recognize unsegmented signs from image streams. Overfitting is one of the most critical  problems in CSLR training, and previous works show that the iterative training scheme can partially solve this problem while also costing more training time. In this study, we revisit the iterative training scheme in recent CSLR works and realize that sufficient training of the feature extractor is critical to solving the overfitting problem. Therefore, we propose a Visual Alignment Constraint (VAC) to enhance the feature extractor with alignment supervision. Specifically, the proposed VAC comprises two auxiliary losses: one focuses on visual features only, and the other enforces prediction alignment between the feature extractor and the alignment module. Moreover, we propose two metrics to reflect overfitting by measuring the prediction inconsistency between the feature extractor and the alignment module. Experimental results on two challenging CSLR datasets show that the proposed VAC makes CSLR networks end-to-end trainable and achieves competitive performance.
\end{abstract}

\section{Introduction}

Sign Language is a complete and natural language that conveys information through both manual components (hand/arm gestures) and non-manual components (facial expressions, head movements, and body postures)~\cite{dreuw2007speech,ong2005automatic} with its own grammar and lexicon~\cite{sandler2006sign}. Vision-based Continuous Sign Language Recognition (CSLR) aims to automatically recognize signs from image streams, which can bridge the communication gap between the Deaf and hearing people. It also provides more non-intrusive communication channel for sign language users. 

\begin{figure}[t]
\begin{center}
\includegraphics[width=0.8\linewidth]{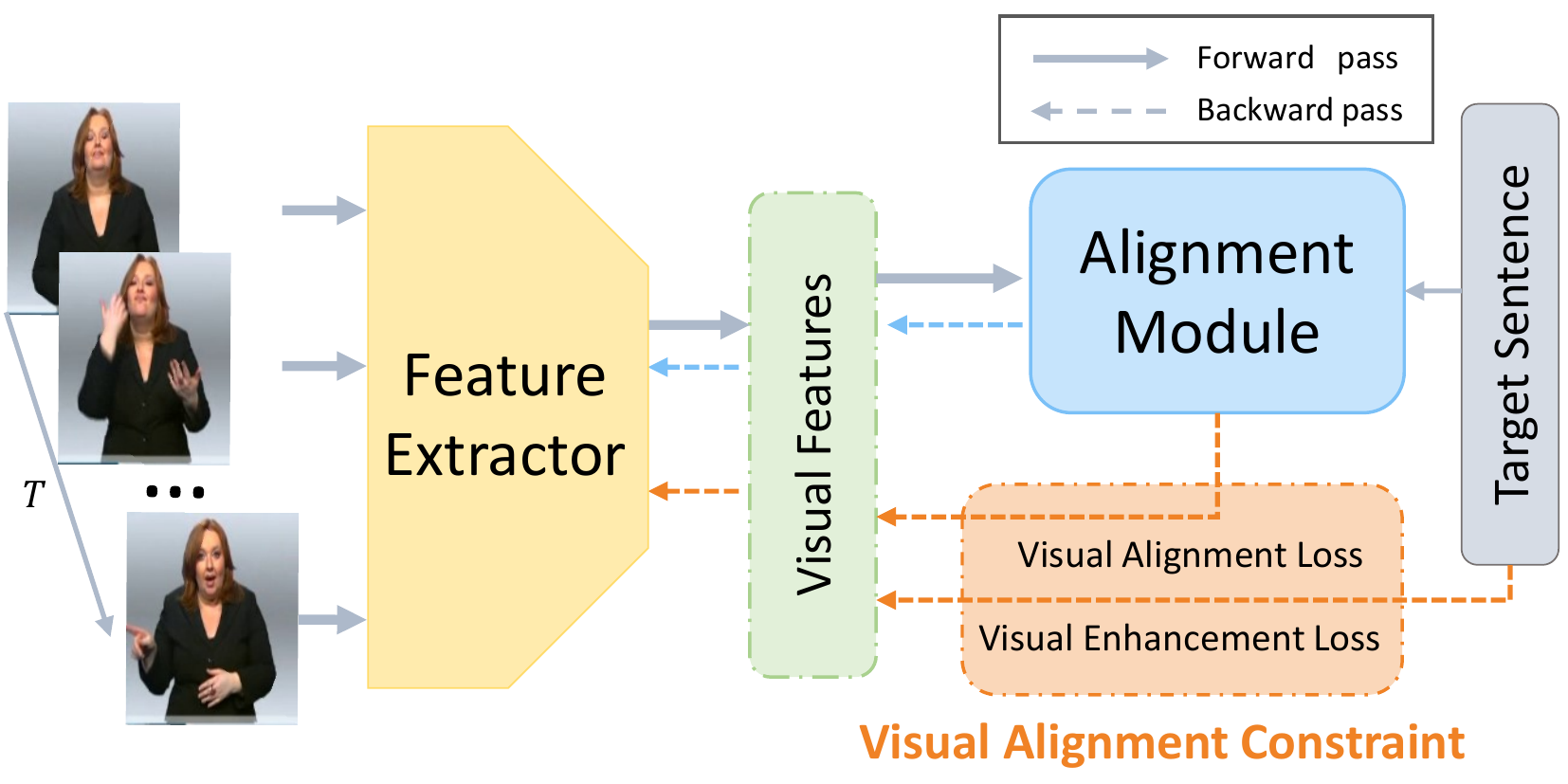}
\end{center}
\vspace{-13pt}
\caption{Overview of the proposed non-iterative CLSR approach with the visual alignment constraint. To solve the insufficient training of the feature extractor, the proposed VAC enhances the generalization ability of the visual extractor by constraining the feature space with the alignment supervision.}
\label{fig:title_figure}
\vspace{-10pt}	
\end{figure}
\smallskip



Different from speech recognition, the data collection and annotation of sign language are costly, which poses a significant problem for recognition~\cite{bragg2019sign}. Therefore, most recent CSLR works solve this problem in a weakly supervised manner and adopt network architectures composed of the feature extractor and the alignment module. The feature extractor abstracts visual information from each frame, and the alignment module searches the possible alignments between visual features and the corresponding labeling. Different to those works~\cite{koller2019weakly,koller2016deephand,koller2017re} adopt HMMs to update frame-wise state labels for the feature extractor, Graves~\etal~\cite{graves2006connectionist} provide a more elegant solution so-called Connectionist Temporal Classification (CTC) to align the prediction and labeling by maximizing the sum of probability of all feasible alignments, which is adopted by following works~\cite{camgoz2017subunets,cheng2020fully,cui2017recurrent,cui2019deep,koller2019weakly,niustochastic,wang2018connectionist}. 

Although CTC-based CSLR methods provide convenience in training, previous studies~\cite{cui2019deep,pu2019iterative} show that end-to-end training limits the discriminative power of the feature extractor. They leverage the iterative training scheme to enhance the feature extractor, which significantly improves the performance.  Nevertheless, it requires an additional fine-tuning process besides the end-to-end training and increases the training time. Several recent works~\cite{cheng2020fully,niustochastic} try to accelerate this training scheme by adopting fully convolutional networks and fine-grained labels. 



In this study, we revisit CTC-based CSLR model at different iterations and observe that only a few frames play key roles in training. The feature extractor abstracts visual information and provides initial localizations of key frames for the alignment module. The alignment module further refines the recognition results from the feature extractor and learns long-term relationships with its powerful temporal modeling ability. Due to the spike phenomenon of CTC~\cite{graves2012supervised,li2020reinterpreting}, the alignment module converges much faster than the feature extractor on CSLR datasets with limited samples and cannot provide enough feedback to the feature extractor. The overfitting of the alignment module leads to insufficient training of the feature extractor and deteriorates the generalization ability of the trained model. The iterative training scheme tries to solve this problem by enhancing the feature extractor with iteratively refined pseudo labels.



Based on above observations, we conclude that constraining the feature space is critical to efficiently train CSLR models. To solve this problem, we propose a Visual Alignment Constraint (VAC) to make CSLR networks end-to-end trainable. As shown in Fig.~\ref{fig:title_figure}, the proposed VAC is composed of two auxiliary losses which provide extra supervision for the feature extractor. The visual enhancement loss enforces the feature extractor to make predictions based on visual features only and the visual alignment loss aligns the short-term visual predictions to long-term contextual predictions. With the combination of the two losses, the proposed method achieves competitive performance to the latest methods on PHOENIX14~\cite{koller2015continuous} and CSL~\cite{huang2018video} datasets.

To better understand the performance gains, we present two metrics named Word Deterioration Rate (WDR) and Word Amelioration Rate (WAR) to evaluate the contributions of the feature extractor and the alignment module, which can also be used as indicators of overfitting. Comparing to the iterative training procedure, experimental results show that the proposed method can obtain a more powerful feature extractor and make better use of visual features.

The major contributions are summarized as follows:
\begin{itemize}
	\item Revisiting the iterative training scheme in CSLR and showing that the overfitting of the alignment module leads to insufficient training of the feature extractor.
	 \item Proposing a visual alignment constraint to make the network end-to-end trainable by enhancing the feature extractor and aligning visual and contextual features.

	 \item Presenting two metrics to evaluate the contributions of the feature extractor and the alignment module, which verifies the effectiveness of the proposed method.
	\end{itemize}


\begin{figure*}[t]
\begin{center}
\includegraphics[width=0.85\linewidth]{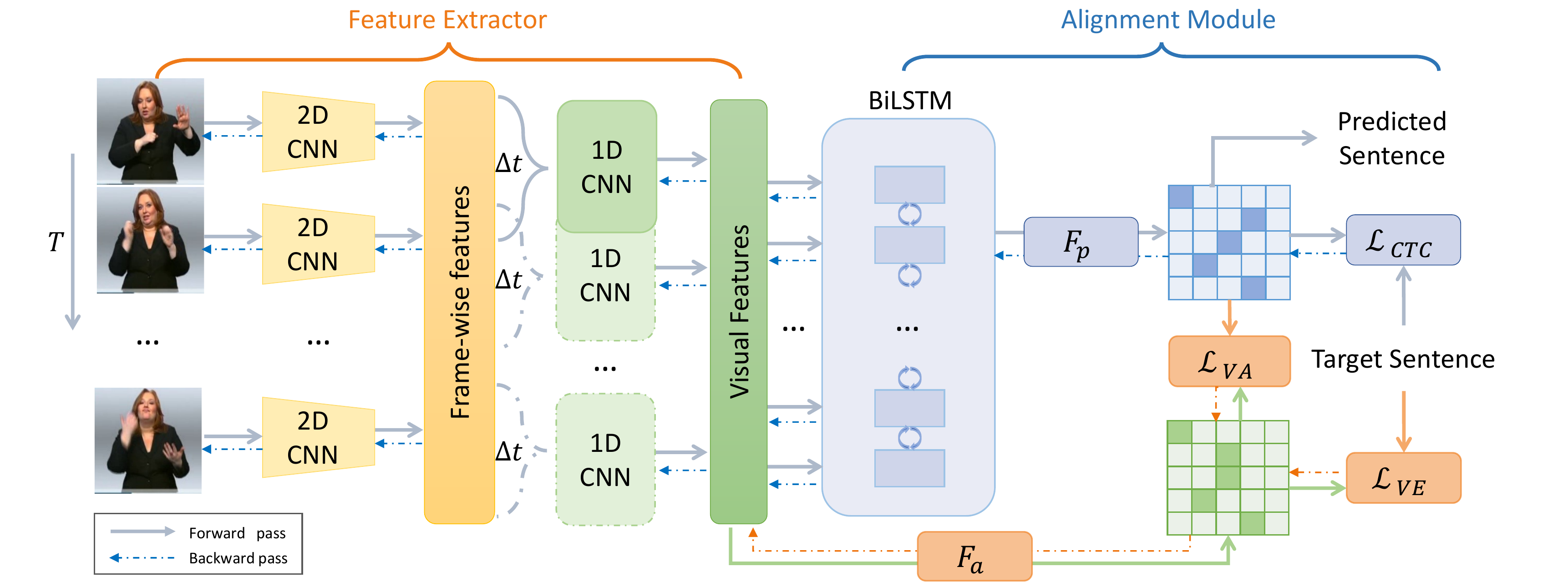}
\end{center}
\vspace{-13pt}
\caption{The proposed framework consists of three components: a feature extractor, an alignment module, and an auxiliary classifier $F_a$. The feature extractor first takes image sequence to abstract frame-wise features, and then applies 1D-CNN to extract the local visual information with $\Delta t$ temporal receptive field. The outputs of 1D-CNN noted as visual features are sent to the alignment model and the auxiliary classifier. Two auxiliary losses are adopted during training: the visual enhancement loss ($\mathcal{L}_{VE}$) aligns visual features and the target sequence, and the visual alignment loss ($\mathcal{L}_{VA}$) aligns short-term visual predictions and long-term context predictions through knowledge distillation.}
\label{fig:architecture}
\vspace{-10pt}
\end{figure*}

\section{Related Work}
\label{sec:related}
\subsection{Continuous Sign Language Recognition}
Sign Language Recognition (SLR) methods can be roughly categorized into isolated SLR~\cite{msasl,li2020word,li2020transferring} and continuous SLR~\cite{cui2019deep,koller2019weakly}. Different to isolated SLR, most CSLR approaches model sequence recognition in a weakly supervised manner: only sentence-level labeling is provided. Some early CSLR methods~\cite{gao2004chinese,han2009modelling,ong2005automatic} adopt a divide-and-conquer paradigm that splits sign video into several subunits with HMM-based recognition systems to work with limited data. Hand-crafted features~\cite{freeman1995orientation,koller2015continuous,sun2013discriminative} are carefully selected to provide better visual information.

The recent successes of CNNs in computer vision~\cite{he2016deep,simonyan2014very,szegedy2015going} provide powerful tools for visual features representation. However, CNNs need frame-wise annotations contrary to the weakly supervised nature of CSLR. To solve this problem, Koller~\etal~\cite{koller2016deephand} propose an iterative expectation-maximization approach by adding a hand shape classifier to the GMM-HMM model as an intermediate task to provide frame-level supervision. A few studies extend this work by proposing CNN+LSTM+HMM framework~\cite{koller2016deep}, incorporating more clues~\cite{koller2019weakly} and improving the iterative alignment approach~\cite{koller2017re}. This iterative CNN-LSTM-HMM setup provides robust visual features that are adopted by many recent works~\cite{camgoz2020sign,cihan2018neural}.

Although the CNN-LSTM-HMM hybrid approaches achieve great results, they still need HMMs to provide frame-wise labels. Graves~\etal~\cite{graves2006connectionist} propose the CTC loss to maximize probabilities of all feasible alignments, which is widely used in many sequence problems~\cite{graves2013speech,graves2008novel}. Several recent works~\cite{camgoz2017subunets,cui2017recurrent} use CTC loss to achieve the end-to-end training of CSLR. However, some works~\cite{cui2017recurrent,cui2019deep,pu2019iterative} find that such an end-to-end approach cannot train feature extractor properly and bring the iterative training back in use. Until very recently, some works~\cite{cheng2020fully,niustochastic} try to solve this problem in an end-to-end way. Cheng~\etal~\cite{cheng2020fully} propose a gloss feature enhancement module to learn better visual features. Niu and Mak~\cite{niustochastic} propose a multiple states approach and several operations to alleviate the overfitting problem. In this work, we try to explore the nature of iterative training and propose a more efficient method to train CSLR models.
\subsection{Auxiliary Learning}
Different from the conventional Multi-Task Learning~\cite{caruana1997multitask}, which aims to improve the generalization of all tasks, auxiliary learning chooses proper auxiliary tasks to assist in the generalization of the primary task. One straightforward way is to combine multiple tasks at the output stage. Follow this idea, Kim~\etal~\cite{kim2017joint} use CTC to speed up the training process and provide a monotonic alignment constraint. Pu~\etal~\cite{pu2019iterative} propose an iteratively alignment network that jointly optimizes the CTC decoder and the LSTM decoder, additionally with a soft-DTW alignment constraint. Goyal~\etal~\cite{NIPS2017_900c563b} propose an auxiliary loss to alleviate the posterior collapsing phenomenon in autoregressive decoder~\cite{bowman2016generating}. Another idea is to use different supervision at different stages. Sanabria~\etal~\cite{sanabria2018hierarchical} use several lower-level tasks, such as phoneme recognition, to constrain intermediate representations for speech recognition. In this study, we adopt the auxiliary learning strategy to provide the visual alignment constraint for the feature extractor.

\section{Revisiting the Iterative Training in CSLR}
\label{sec:revisit}
The CSLR aims to predict the corresponding gloss label sequence $\boldsymbol{l}=(l_1,\cdots,l_N)$ based on a sequence of $T$ frames $\boldsymbol{X}=(\boldsymbol{x}_1,\cdots,\boldsymbol{x}_T)$. The feature extractor plays an important role in CSLR, which extracts visual features $\boldsymbol{V}=(\boldsymbol{v}_1,\cdots,\boldsymbol{v}_{T'})$ from image sequences. As shown in Fig.~\ref{fig:architecture}, we choose 2D-CNN to extract frame-wise features and 1D-CNN to extract local posture and motion information from neighboring frames as previous works did~\cite{cui2019deep,zhou2020spatial}. The gloss-wise features are fed into a two-layer BiLSTM and the primary classifier $F_p$ to combine long-term relationships and provide the predicted logits $\boldsymbol{Z}=(\boldsymbol{z}_1,\cdots,\boldsymbol{z}_{T'})$. CTC loss is adopted to provide supervision by aligning the predictions and sequence labelings.


\begin{figure*}[t]
\subfigure[Iteration 1]{
\begin{minipage}[t]{0.348837\linewidth}
\begin{center}
\includegraphics[width=0.9\linewidth,page=1]{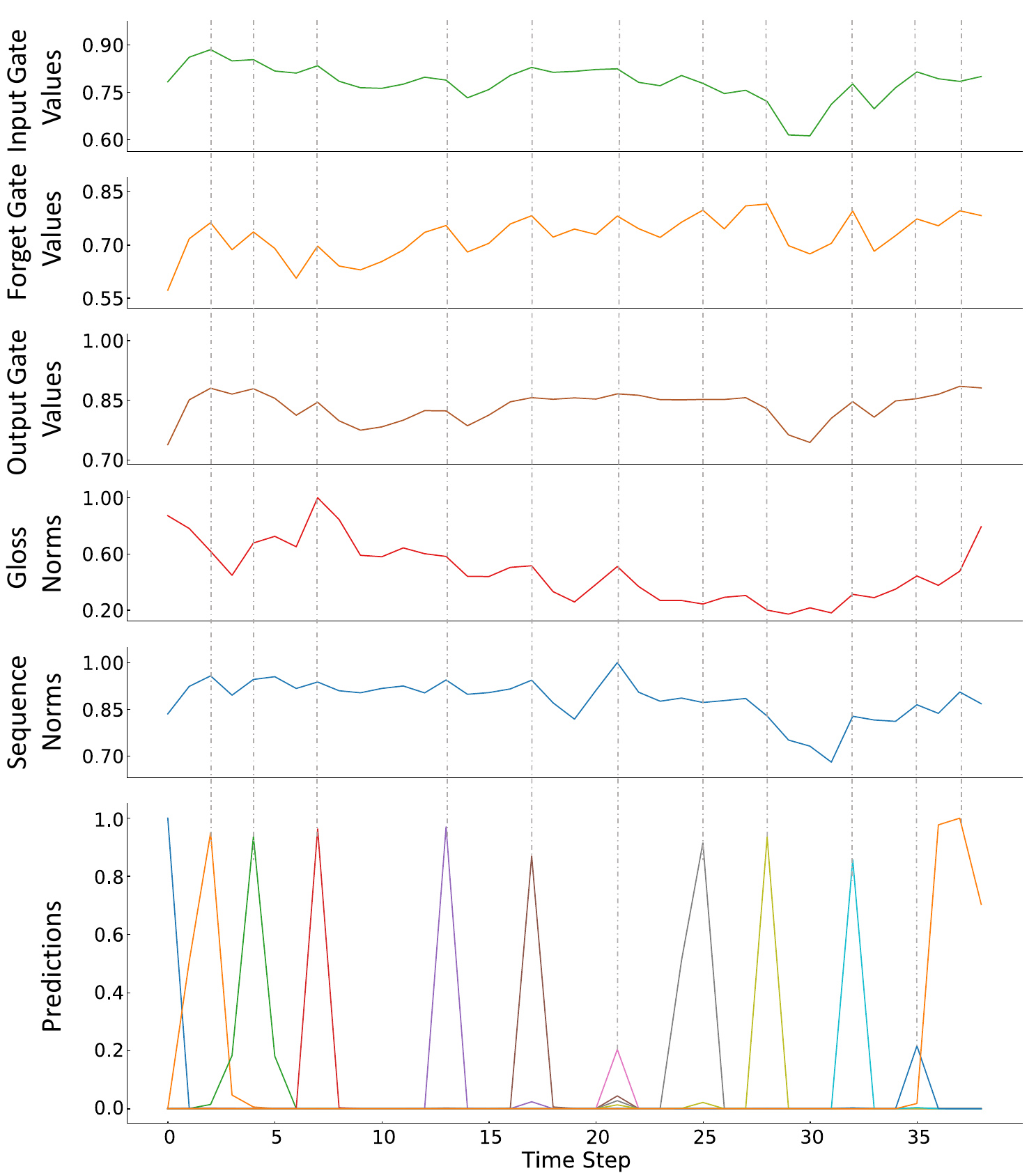}
\end{center}
\vspace{-10pt}
\label{fig:iter0}
\end{minipage}%
}%
\subfigure[Iteration 2]{
\begin{minipage}[t]{0.32558\linewidth}
\begin{center}
\includegraphics[width=0.9\linewidth,page=2]{imgs/figure3_iterative_training.pdf}
\end{center}
\vspace{-10pt}
\label{fig:iter1}
\end{minipage}
}%
\subfigure[Iteration 3]{
\begin{minipage}[t]{0.32558\linewidth}
\begin{center}
\includegraphics[width=0.9\linewidth,page=3]{imgs/figure3_iterative_training.pdf}
\end{center}
\vspace{-10pt}
\label{fig:iter2}
\end{minipage}
}%
\label{fig:pointlstm}
\caption{Visualization of the gate values, the $l_2$ norm of features and the final prediction of a training sample among different iterations.}
\vspace{-10pt}
\label{fig:iter}
\end{figure*}

\subsection{The Spike Phenomenon of CTC}
\label{sec:spike}
The Connectionist Temporal Classification~\cite{graves2006connectionist} is designed for end-to-end temporal classification tasks with unsegmented data. To provide more effective supervision, CTC introduces a `blank' to represent unlabeled data (such as movement epenthesis or non-gesture segments in CSLR) and solves the alignment problem with dynamic programming. The blank class and gloss vocabulary $\mathbb{G}$ build the final extended gloss vocabulary $\mathbb{G}'=\mathbb{G}\cup\{blank\}$.


CTC defines a many-to-one function $\mathcal{B}:\mathbb{G}'^T\rightarrow \mathbb{G}^{\leq T}$ to align label sequence referred to as path $\pi\in \mathbb{G}'^{T}$ and labeling $\boldsymbol{l}\in \mathbb{G}^{\leq T}$ by sequentially removing the repeated labels and the blanks from the path. For example, $\mathcal{B}(\text{-}aaa\text{-}\text{-}aabbb\text{-})=\mathcal{B}(\text{-}a\text{-}ab\text{-})=aab$. With the help of this function, CTC can provide supervision for parameters $\theta$ of the feature extractor and the alignment module by summing the probabilities of all feasible paths:
\begin{equation}
\begin{aligned}
\mathcal{L}_{CTC}&=-\log p(\boldsymbol{l}|\boldsymbol{X};\theta) \\
&=-\log\Big(\sum_{\pi\in \mathcal{B}^{-1}(\boldsymbol{l})} p(\pi|\boldsymbol{X};\theta)\Big).
\end{aligned}
\end{equation}
The conditional probability $p(\pi|X)$ can be calculated according to the conditional independence assumption:
\begin{equation}
\begin{aligned}
p(\pi|\boldsymbol{X})=\prod_{t=1}^{T'} p(\pi_t|\boldsymbol{X};\theta), \\
\end{aligned}
\end{equation}
where the probabilities are calculated by applying softmax fuction to the the network output logits: $P_\theta=\text{softmax}(\boldsymbol{Z})$.	


As mentioned above, CTC aligns the path and the labeling by introducing a blank class and removing the repeat labels. When optimizing network with CTC, predictions tend to form a series of spike responses~\cite{graves2006connectionist,li2020reinterpreting}. The main reason for this is that predicting a blank label is a much safer choice for CTC when the network cannot confidently distinguish gloss boundaries. For example, both $\mathcal{B}(aaab)$ and $\mathcal{B}(a \text{-}\text{-} b)$ are corresponding to the same labeling, but $\mathcal{B}(a\bm{b}ab)$ will bring larger loss even if there is only one mistake. 
Therefore, the CTC loss mainly focuses on key frames, and the final predictions are composed of a few non-blank key frames and many high-confidence blank frames.



\subsection{Visualization of LSTM Gates}
Long Short-Term Memory~\cite{hochreiter1997long} is widely used in sequence modeling, which excellently models long-term dependencies. The core component of LSTM is its memory design: the input and forget gates control information from current inputs and the past memory to the current memory. The output gate controls what is expected to output from the current memory. The total update mechanism is as follows ( $\odot$ denotes the Hadamard product):
\begin{equation}
\begin{aligned}
\boldsymbol{i}_t &=\sigma(\boldsymbol{U}_i\boldsymbol{v}_t+\boldsymbol{W}_i\boldsymbol{h}_{t-1}+\boldsymbol{b}_i),\\
\boldsymbol{f}_t &=\sigma(\boldsymbol{U}_f\boldsymbol{v}_t+\boldsymbol{W}_f\boldsymbol{h}_{t-1}+\boldsymbol{b}_f),\\
\boldsymbol{o}_t &=\sigma(\boldsymbol{U}_o\boldsymbol{v}_t+\boldsymbol{W}_o\boldsymbol{h}_{t-1}+\boldsymbol{b}_o),\\
\tilde{\boldsymbol{c}_t} &=\sigma(\boldsymbol{U}_c\boldsymbol{v}_t+\boldsymbol{W}_c\boldsymbol{h}_{t-1}+\boldsymbol{b}_c),\\
\boldsymbol{c}_t &= \boldsymbol{f}_t\odot \boldsymbol{c}_{t-1}+\boldsymbol{i}_t\odot \tilde{\boldsymbol{c}}_t,\\
\boldsymbol{h}_t &= \boldsymbol{o}_t\odot \tanh(\boldsymbol{c}_t).
\label{equ:lstm}
\end{aligned}
\end{equation}

Here the $i_t,f_t$ and $o_t$ are corresponding to input, forget and output gates, respectively, the vector $h_t$ and $c_t$ are hidden and cell states. where $\boldsymbol{U}_\cdot$ and $\boldsymbol{W}_\cdot$ are the input-to-hidden and hidden-to-hidden weight matrices, and $\boldsymbol{b}_\cdot$ are bias vectors. Element-wise sigmoid is reprensented by $\sigma$.

Previous works~\cite{cui2017recurrent,cui2019deep,pu2019iterative} adopt iterative training to enhance the visual extractor. To explore how iterative training works and how LSTM makes predictions in CSLR, we begin by visualizing the averaged gate values of the last forward-direction LSTM and the network predictions at different iterations in Fig.~\ref{fig:iter}. For the predictions, we only visualize non-blank classes that occur in the labeling.  We can make some observations from the comparison of line charts:

%



1) The gate values and the predictions have positive correlations on the training set, and they reach the local maximum on similar frame subsets.

2) The correlations appear to be weakened as the iteration progresses, especially for the input and output gates, which become larger and smoother.

The above two observations are quite puzzling, as three gates are expected to play different roles in information flow. As shown in Equ.~\ref{equ:lstm}, three gates take the same inputs and have independent parameters. Therefore, we pinpoint the problem to the magnitude of input features and further visualize the $l_2$ norms of the activations before the first and the second BiLSTM layers, which are referred to as the gloss and sequence norms in Fig.~\ref{fig:iter}.

\subsection{A Magnitude Hypothesis}
\label{sec:hypo}

Fig.~\ref{fig:iter} presents an interesting observation that the $l_2$ norms of gloss and sequence features have similar tendencies with gates values and final predictions. Besides, the magnitudes variances of both gloss and sequence become smaller as the iteration progresses.  Several recent papers~\cite{li2020overcoming,wang2017normface} found that well-separated features tend to have larger magnitudes, and we hypothesize the magnitudes variances are relevant to the importance of frames:

\emph{The $l_2$ norms of the features are effect indicators that reflect frame importance: the optimization algorithm will decrease the magnitudes of activations when suppressing the non-key frames due to the spike phenomenon of CTC.}



With the above hypothesis, it is clear that frames with larger magnitudes in Fig.~\ref{fig:iter} play key roles compared to their neighbors. We further interpret the learning process of CTC-based CSLR model into two stages: 1) the feature extractor provides visual and initial localization information for the alignment module, and 2) the BiLSTM layers refine the localization and learn long-term relationships among key frames. Such a learning scheme can make efficient use of the data and accelerate the training process. 

However, current CSLR datasets contain less data than other sequence learning tasks~\cite{graves2013speech,hannun2014deep}, which means the BiLSTM layers can easily overfit the whole training set with partial visual information and other frames are decreasingly involved in the training progress. Although the network can achieve stable convergence, the power of feature extractor is not sufficiently explored. Therefore, the feature extractor cannot provide robust visual features during inference and deteriorate the generalization performance.

Based on these analyses, we attribute the success of iterative training to the reduction of the overfitting problem. With pseudo labels generated by the alignment module, the fine-tuning stage can enhance the feature extractor to make it generalize better. Although the pseudo labels can relieve the overfitting problem in some sense, it is still not enough. Therefore, we propose the visual alignment constraint on the visual feature space, which enforces the feature extractor to make predictions on its own and adopts the distillation loss to align both visual and contextual spike responses.

\section{Visual Alignment Constraint}
\label{sec:vac}
As mentioned above, the BiLSTM layers can easily overfit the training set with partial visual information. In this paper, we propose the Visual Alignment Constraint (VAC) to enhance the feature extractor with more alignment supervision. The proposed VAC is implemented by two simple auxiliary losses: the Visual Enhancement (VE) loss and the Visual Alignment (VA) loss. Besides, we propose two new evaluation metrics, Word Deterioration Rate (WDR) and Word Amelioration Rate (WAR), to evaluate the contributions of the feature extractor and the alignment module. 





\subsection{Loss Design of VAC}

\noindent\textbf{VE Loss.}
To enhance the feature extractor, we proposed to add an auxiliary classifier $F_a$ on visual features $\boldsymbol{V}$ to get the auxiliary logits $\tilde{\boldsymbol{Z}}=(\tilde{\boldsymbol{z}}_1,\cdots,\tilde{\boldsymbol{z}}_T')=F_a(\boldsymbol{V})$ and propose the VE loss that directly provides supervision for the feature extractor. This auxiliary loss enforces the feature extractor to make predictions based on local visual information only. Compared to previous gloss-wise supervision that needs to generate pseudo labels, we propose to add a CTC loss on the auxiliary classifier as the VE loss, which is compatible with the primary CTC loss and flexible to network designs. The VE loss only provides supervision for parameters $\theta^v$ of the feature extractor and the auxiliary classifier:

\begin{equation}
\mathcal{L}_{VE}=\mathcal{L}_{CTC}^v = -\log p(\boldsymbol{l}|\boldsymbol{X};\theta^v).
\end{equation}

\noindent\textbf{VA Loss.}
Because the VE loss lacks contextual information and is independent of the primary loss, which may lead to misalignment between two classifiers, we further propose the VA loss. The VA loss is implemented as a knowledge distillation loss~\cite{distillation}, which regards the entire network and the visual feature extractor as the teacher and student models, respectively. A high temperature $\tau$ is adopted to ``soften'' probability distribution from spike responses. The distillation process is formulated as: 
\begin{equation}
\begin{aligned}
\mathcal{L}_{VA} &= \text{KL}\Big(\text{softmax}(\frac{\boldsymbol{Z}}{\tau}),\text{softmax}(\frac{\tilde{\boldsymbol{Z}}}{\tau})\Big).
\end{aligned}
\end{equation}

In summary, to achieve the visual alignment goal, the VE loss enforces the feature extractor to provide more robust visual features for the alignment module, while the VA loss aligns the predictions of two classifiers by providing long-term supervision for the visual extractor. With the help of both losses, the feature extractor obtains more supervision which is compatible with the alignment module. The final objective function is composed of the primary CTC loss, the visual enhancement loss, and the visual alignment loss:

\begin{equation}
\mathcal{L} = \mathcal{L}_{CTC} + \mathcal{L}_{VE} + \alpha\mathcal{L}_{VA}.
\end{equation}

\subsection{Prediction Inconsistency Measurement}

Word Error Rate (WER) is a widely-used metric to evaluate the performance of recognition algorithms in CSLR~\cite{koller2015continuous}. It is also referred to as the length normalized edit distance, which first aligns the recognized sequence with the reference sentence and then counts the number of operations, including substitution (sub), deletion (del), and insertion (ins), to transfer from the reference to the recognized sequence: \text{WER} = (\text{\#sub} + \text{\#del} + \text{\#ins}) / \text{\#reference}.

As shown in Fig.~\ref{fig:war}, both of the auxiliary and the primary recognized sentences ($\text{HYP}_a$ and $\text{HYP}_p$) have the same WER 22.22\% ($\text{HYP}_a$ has two deletion errors, and $\text{HYP}_p$ has two insertion errors). The primary classifier corrects the misrecognized results of the auxiliary classifier but makes new mistakes, which can not be measured by WER. Therefore, we firstly align sentence triplet ($\text{REF}^*, \text{HYP}^*_a, \text{HYP}^*_p$) and then calculate WDR and WAR: WDR measures the ratio that is correctly recognized by the auxiliary classifier but misrecognized by the primary classifier (two `SUED' in $\text{HYP}^*_p$), and WAR does in the opposite direction (`MEHR' and `KALT' in $\text{HYP}^*_p$). With the proposed metrics, we can connect the $\text{WER}^*$\footnote {The adopted alignment approach leads to a little performance degradation than the general WER.} performance of two classifiers by:
\begin{equation}
\text{WER}_p^* =  \text{WER}_a^* + \text{WDR} - \text{WAR}.
\label{equ:wer}
\end{equation}

\begin{figure}[h]
\centering
\vspace{-15 pt}
\includegraphics[width=1.0\linewidth,page=1]{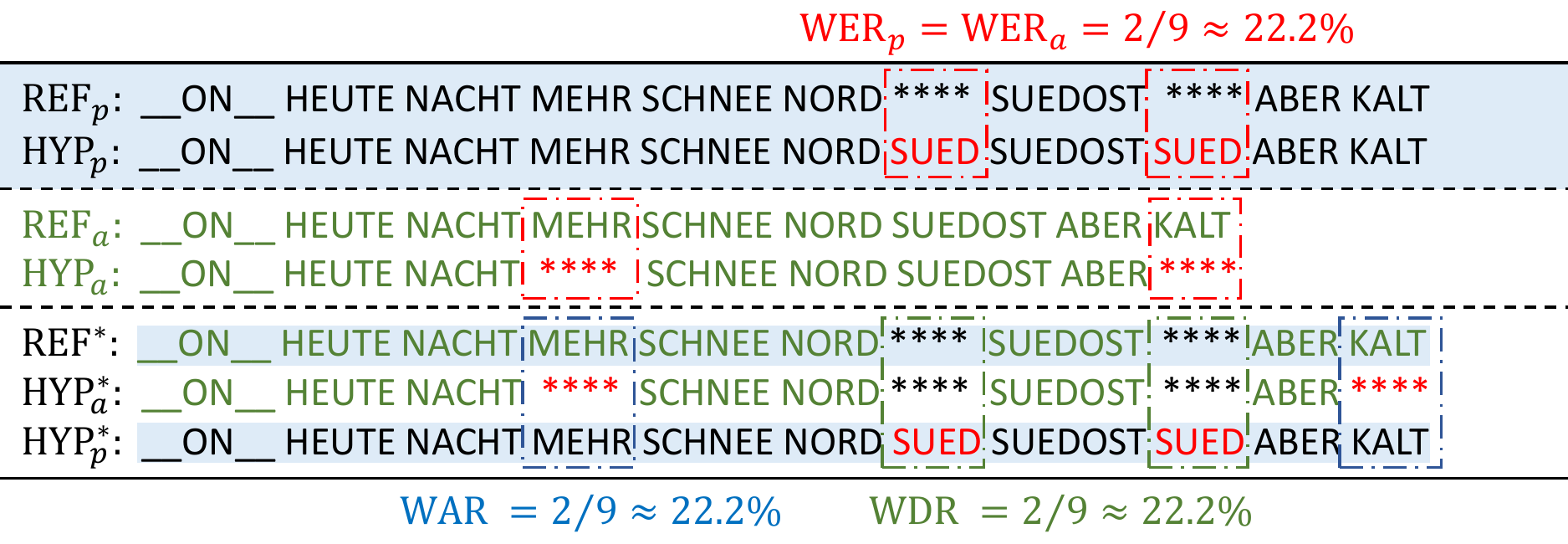}
\label{fig:war}
\vspace{-10 pt}
\caption{ Alignment results of the proposed metrics. We highlight {\color{red}{wrong recognized glosses}} and the alignment results of \textcolor[rgb]{0.3254,0.5098,0.2078}{the auxiliary classifier} and \colorbox[rgb]{0.8705,0.9216,0.9686}{\color{black}the primary classifier}.} 
\label{fig:war}
\end{figure}

In Equ.~\ref{equ:wer}, the final result $\text{WER}_p^*$ come from three aspects: how well the visual extractor performs (related to $\text{WER}_a^*$), how much visual information is not fully utilized (related to $\text{WDR}$) and how many predictions are made by contextual information only (related to $\text{WAR}$). More details are given in the supplementary material.

\section{Experiments}
\label{sec:exp}
\subsection{Experimental Setup}
\noindent\textbf{Datasets.} We evaluate the proposed method on two widely used datasets: RWTH-PHOENIX-Weather-2014 (PHOENIX14)~\cite{koller2015continuous} and Chinese Sign Language (CSL) dataset~\cite{huang2018video}.  All ablations are performed on PHOENIX14.

The PHOENIX14 dataset is a widely used CSLR dataset recorded from the German TV weather forecasts and performed by nine hearing SL interpreters. It contains 6841 sentences with 1295 different glosses.  The dataset is split into 5672 training sentences, 540 development (Dev) sentences, and 629 test sentences for the multi-signer setup.

The CSL dataset is collected under laboratory conditions with 100 sign language sentences with a vocabulary size of 178. Fifty signers perform each sentence five times (in 25000 videos with 100+ hours). We follow the previous setting~\cite{cheng2020fully} and split the dataset into training and test sets according to the ratio of 8:2.




\noindent\textbf{Implementation Details.}
ResNet18~\cite{he2016deep} is picked as the frame-wise feature extraction in considering its efficiency on the PHOENIX14 dataset. For the CSL dataset, we adopt VGG11~\cite{simonyan2014very} as the backbone to reduce side effects of inconsistent statistics under the signer-independent setting. The gloss-wise temporal layer and two BiLSTM layers with 2$\times$512 dimensional hidden states are adopted as the default setting. The weight $\alpha$ for $\mathcal{L}_{VA}$ is set to 25 and its temperature $\tau$ is set to 8 by default. We train all the models for 80 epochs for PHOENIX14 and 20 epochs for CSL with a mini-batch size of 2. Adam optimizer is used with an initial learning rate of $10^{-4}$, divided by five after 40 and 60 epochs for PHOENIX14 and 10 and 15 for CSL. For iterative training, we reduce the learning rate by a factor of five after each iteration.  All frames are resized to 256x256, and the training set is augmented with random crop (224x224), horizontal flip ($50\%$), and random temporal scaling ($\pm20\%$).

\begin{table}[]
\centering
\setlength{\tabcolsep}{12pt}
\caption{ Ablation results (WER, \%) of iterative training and BN.}
\begin{tabular}{ccccc}\shline
\multirow{2}{*}{Iterations} & \multicolumn{2}{c}{w/o BN} & \multicolumn{2}{c}{w/ \ \; BN} \\ \cline{2-5} 
 & Dev & Test & Dev & Test \\\hline
1 & 32.7 & 33.0 & 27.2 & 28.0 \\
2 & 28.9 & 29.8 & 25.5 & 26.3 \\
3 & 28.3 & 28.9 & \textbf{24.7} & \textbf{26.2} \\ \hline
None & 30.4 & 32.1 & 25.4 & 26.6 \\ \shline
\end{tabular}
\label{tab: iter}
\vspace{-10pt}
\end{table}



\begin{table}[]
\centering
\setlength{\tabcolsep}{8.5pt}
\caption{Ablation results (WER, \%) of Learning Rate (LR) ratios (LR of the feature extractor / LR of the alignment model).}

\begin{tabular}{cccccc}\shline
LR Ratio & 0.1 & 0.5 & 1 & 2 & 10 \\ \hline
Dev & \textbf{25.0} & 25.6 & 25.4 & 26.9 & 34.8 \\
Test & \textbf{25.6} & 26.5 & 26.6 & 27.5 & 35.1 \\ \shline
\end{tabular}
\label{tab: lr}
\end{table}



\begin{table}[!t]
\centering
\caption{ Ablation results (WER,\%) of VAC design.}
\begin{tabular}{lccccc}
\shline
 & $\mathcal{L}_{CTC}$ & $\mathcal{L}_{VE}$ & $\mathcal{L}_{VA}$ & Dev & Test \\ \hline
Baseline & \checkmark &  &  & 25.4 & 26.6 \\
Baseline+VE & \checkmark & \checkmark &  & 23.3 & 23.8 \\
Baseline+VA & \checkmark &  & \checkmark & 24.5 & 25.1 \\
Baseline+VAC & \checkmark & \checkmark & \checkmark & \textbf{21.2} & \textbf{22.3}\\
\shline
\end{tabular}
\vspace{-10pt}
\label{tab:aux}
\end{table}

\subsection{Quantitative Results}

\noindent\textbf{Ablation on iterative training and BN.} Batch Normalization (BN)~\cite{ioffe2015batch} is a widely-used tool to accelerate the training of deep networks by normalizing the activations. Although we adopt a small batch size, BN significantly improves the performance. As shown in Table~\ref{tab: iter}, adding a BN layer after each temporal convolution layer brings 5.5\%, 3.4\%, and 3.6\% performance gains at each iteration on the Dev set, which indicates the existence of insufficient training of the feature extractor. We can also observe that adopting iterative training can lead to noticeable performance gains compared to non-iterative training. 

\noindent\textbf{Ablation on learning pace.} A natural idea to solve the insufficient training problem is adjusting the learning paces of the feature extractor and the alignment module. In Table~\ref{tab: lr}, we compare results under different learning rate ratios. Adopting a smaller learning rate for the feature extractor leads to comparable results with iterative training, which suggests the existence of insufficient training. However, it is hard to find an optimal learning setting. We adopt a non-iterative model with BN layers and the normal 1:1 learning rate ratio as our baseline.

\begin{figure*}[t]
\begin{center}
\includegraphics[width=0.85\linewidth]{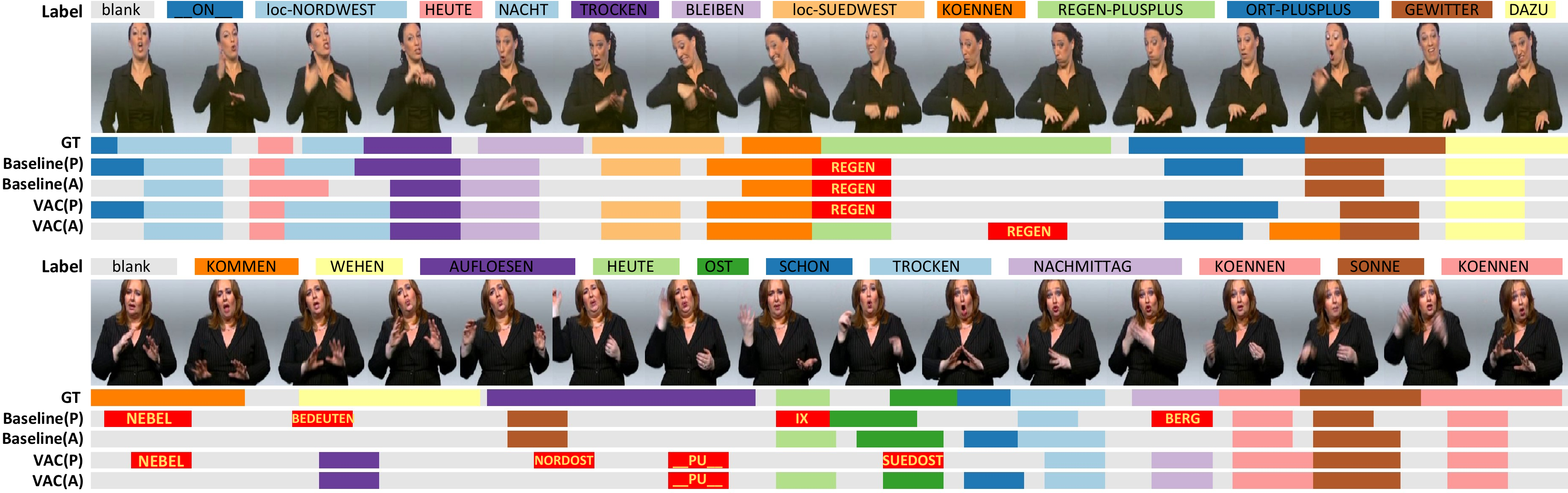}
\end{center}
\vspace{-10pt}
\caption{Qualitative comparison among different settings with examples from training (the upper) and Dev (the lower) sets of PHOENIX14. Wrong recognized glosses (except del) are marked in red. The primary classifier and auxiliary classifier outputs are marked as (P) and (A).}
\label{fig:example}
\vspace{-15pt}
\end{figure*}

\begin{figure}[t]
\subfigure[Results on PHOENIX14 \textbf{training} set.]{
\begin{minipage}[t]{1\linewidth}
\begin{center}
\includegraphics[width=0.9\linewidth,page=2]{imgs/figure5_war_wer_img.pdf}
\end{center}
\vspace{-10pt}
\label{fig:wer1}
\end{minipage}%
}%
\vspace{-5pt}

\subfigure[Results on PHOENIX14 \textbf{Dev} set.]{
\begin{minipage}[t]{1\linewidth}
\begin{center}
\includegraphics[width=0.9\linewidth,page=1]{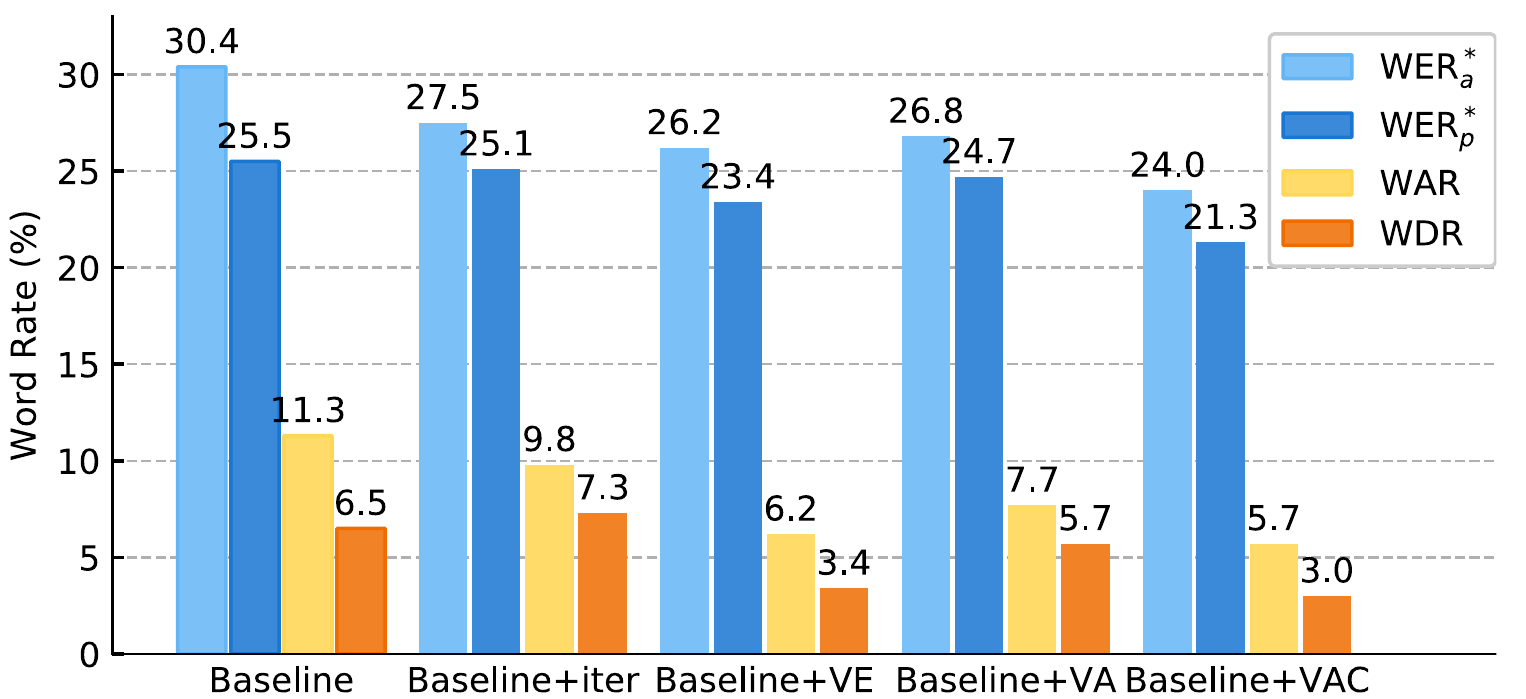}
\end{center}
\vspace{-10pt}
\label{fig:wer2}
\end{minipage}
}%
\label{fig:pointlstm}
\caption{Performance comparison with different metrics and settings ($\Delta \text{WER}^*$ = $\text{WER}^*_a$$-$$\text{WER}^*_p$ = $\text{WAR}$$-$$\text{WDR}$).}
\vspace{-10pt}
\label{fig:wer}
\end{figure}

\noindent\textbf{Ablation on VAC.} Ablations on VAC are presented in Table~\ref{tab:aux}. Constraining visual features with $\mathcal{L}_{VE}$ and $\mathcal{L}_{VA}$ improves the recognition results (2.1\% and 0.9\% on Dev set), which verifies the need to strengthen supervision on the feature extractor. It is also worth noting that although adopting the $\mathcal{L}_{VA}$ only leads to smaller gains than the $\mathcal{L}_{VE}$ only, adopting both losses can achieve further improvement. It suggests that aligning two spike responses provides more effective supervision than adopting independent supervision or distillation only.

\noindent\textbf{Obeservations about the overfitting problem.} Fig.~\ref{fig:wer} visualizes performance comparison with different evaluation metrics and we can draw some interesting observations about overfitting. First, the primary classifier can reach much lower WER on the training set than the auxiliary classifier in Fig.~\ref{fig:wer1}, which reflects its powerful temporal modeling ability. Second, there exists a significant performance gap between the training and Dev sets on WDR, which indicates that the BiLSTM layers do not fully incorporate the visual information although it successfully overfits the training set. Third, the actual performance gap is much larger than WER shows ($\Delta \text{WER}^*$). For example, the performance gap between two classifiers of Baseline on Dev set in Fig.~\ref{fig:wer2} is only \textbf{4.9\%} (=30.4\%-25.5\%), however, the primary classifier makes \textbf{11.3\%} correct predictions based on contextual information only ($\text{WAR}$) and ignores \textbf{6.5\%} correct visual information ($\text{WDR}$). The proposed inconsistent prediction metrics provide a helpful tool to understand and evaluate the overfitting problem.


\label{sec:overfitting}
\noindent\textbf{Obeservations about the performance gap.} Another interesting observation from Fig.~\ref{fig:wer2} is that while the iterative training strengthens the visual extractor, it also increases the $\text{WDR}$. We assume that the pseudo-label-based approach is not well compatible with the primary CTC loss (previous work~\cite{cheng2020fully} adopts a balanced ratio to reduce the effects of ``blank'' labels). Therefore, we adopt an additional CTC loss as our $\mathcal{L}_{VE}$ and it significantly improves both $\text{WAR}$ and $\text{WDR}$. The proposed $\mathcal{L}_{VA}$ has a limited effect on the visual extractor but it can narrow the performance gap between two classifiers. The combined use of both auxiliary losses achieves better performance with a smaller actual performance gap ($\text{WDR}$ and $\text{WAR}$), which verifies the effectiveness of the proposed visual alignment constraint.


\begin{table}[]
\centering
\caption{Ablation results (WER, \%) of temporal layer design. C$\beta$ and P$\beta$ correspond to 1D convolutional layer and max pooling layer with a kernel size of $\beta$, respectively.}
\begin{tabular}{llcc}
\shline
 					 & Temporal Layers   & $\Delta t$ & Dev / Test \\\hline
\multirow{2}{*}{Frame-wise}           & C1                                 & 1 &   25.2 / 26.5                    \\
&C3 & 3 & 24.4 / 25.4\\
Subgloss-wise        & C5-P2                             & 6 &  24.0 / 24.3  \\
Gloss-wise           & C5-P2-C5-P2                       & 16 &   \textbf{21.2} / \textbf{22.3}      \\\shline            
\end{tabular}
\label{tab: arch}
\vspace{-10pt}
\end{table}

\begin{table*}[ht]
\centering
\setlength{\tabcolsep}{11.5pt}
\caption{Performance comparison on PHOENIX14 dataset. Results of the proposed method are based on ResNet18 and Gloss-wise temporal layer. The entries denoted by ``*'' used extra clues (such as keypoints and tracked face regions).}
\begin{tabular}{cccccccc}\shline
\multirow{2}{*}{Methods} & \multirow{2}{*}{Backbone} & \multirow{2}{*}{Iteration}  & \multicolumn{2}{c}{Dev(\%)} & \multicolumn{2}{c}{Test(\%)} \\
 &  &  &   del/ins & WER & del/ins & WER \\\hline
SubUNet~\cite{camgoz2017subunets} & CaffeNet &   & 14.6/4.0 & 40.8 & 14.3/4.0 & 40.7 \\
Staged-Opt~\cite{cui2017recurrent} & VGG-S/GoogLeNet & \checkmark  & 13.7/7.3 & 39.4 & 12.2/7.5 & 38.7 \\
Align-iOpt~\cite{pu2019iterative} & 3D-ResNet & \checkmark  & 12.6/2.6 & 37.1 & 13.0/2.5 & 36.7 \\
Re-Sign~\cite{koller2017re} & GoogLeNet & \checkmark  & - & 27.1 & - & 26.8 \\
SFL~\cite{niustochastic} & ResNet18 &   & 7.9/6.5 & 26.2 & 7.5/6.3 & 26.8 \\
STMC~\cite{zhou2020spatial} & VGG11 & \checkmark & - & 25.0 & - & - \\
DNF~\cite{cui2019deep} & GoogLeNet & \checkmark  & 7.8/3.5 & 23.8 & 7.8/3.4 & 24.4 \\
FCN~\cite{cheng2020fully} & Custom &   & - & 23.7 & - & 23.9 \\
CMA~\cite{pu2020boosting} & GoogLeNet & \checkmark  & 7.3/2.7 & \textbf{21.3} & 7.3/2.4 & \textbf{21.9} \\
\hline
CNN+LSTM+HMM~\cite{koller2019weakly}* & GoogLeNet & \checkmark  & - & 26.0 & - & 26.0 \\
DNF~\cite{cui2019deep}* & GoogLeNet & \checkmark  & 7.3/3.3 & 23.1 & 6.7/3.3 & 22.9 \\
STMC~\cite{zhou2020spatial}* & VGG11 & \checkmark  & 7.7/3.4 & \textbf{21.1} & 7.4/2.6 & \textbf{20.7} \\
\hline
Baseline & ResNet18 &  &  8.3/3.1 & 25.4 & 8.8/3.2 & 26.6 \\
Baseline+VAC & ResNet18 &  &  7.9/2.5 & \textbf{21.2} & 8.4/2.6 & \textbf{22.3} \\\shline
\end{tabular}
\label{tab: phoenix}
\vspace{-10pt}
\end{table*}


\noindent\textbf{Ablation on temporal network design.} Previous pseudo-label-based methods need to carefully design the temporal receptive field, which is set to approximate the average length of the isolated sign~\cite{cheng2020fully,cui2019deep}. Table~\ref{tab: arch} presents the performance comparison with different temporal receptive fields $\Delta t$ to show the effectiveness and flexibility of the proposed VAC. To our surprise, the frame-wise feature extractor still achieves competitive results to other settings, and there is a small performance differences in the temporal layer design. The VAC provides more flexible supervision for the feature extractor and results show that it is superior to iterative training sceme~\cite{cui2019deep}.





\subsection{Qualitative Results}

\noindent\textbf{Results Visualization.}
To better understand the learning process, we give some recognized examples in Fig.~\ref{fig:example}. The upper sample from \textbf{the training set} shows that the auxiliary classifier of the baseline does not correctly recognize some glosses (\textcolor[rgb]{0.4471,0.6078,0.6863}{NACHT}, \textcolor[rgb]{0.8922,0.5451,0.0}{loc-SUEDWEST}, \textcolor[rgb]{0.1216,0.4706,0.7059}{ORT-PLUSPLUS }), but the primary classifier can still deliver the correct result. Although it is reasonable for the primary classifier to make predictions based on contextual information only, the lack of constraint on the feature space increases the risk of overfitting, which may lead to unpredictable predictions when context changes during inference. With the help of the VAC, both auxiliary and primary classifiers are sufficiently trained and make better predictions on the training set.

The lower sample from \textbf{the Dev set} shows a failure case of the alignment module. The auxiliary classifier makes the correct predictions (\textcolor[rgb]{0.4980,0.6745,0.3412}{HEUTE}, \textcolor[rgb]{0.2,0.6275,0.1686}{OST} and \textcolor[rgb]{0.1216,0.4706,0.7059}{SCHON}) based on visual features only. Nevertheless, the primary classifier neglects this information and gives a worse result, which is not mentioned in the WER metric but can be identified by the proposed metrics. More qualitative results can be found in the supplementary material.


\begin{table}[t]
\centering
\setlength{\tabcolsep}{14pt}
\caption{Performance comparison (\%) on CSL dataset. The entry denoted by ``*'' used extra clues (keypoints).}
\begin{tabular}{cccccccc}\shline
Methods & WER \\\hline
LS-HAN~\cite{huang2018video} & 17.3\\
SubUNet~\cite{camgoz2017subunets} & 11.0\\
SF-Net~\cite{yang2019sf} & 3.8 \\
FCN~\cite{cheng2020fully} & 3.0\\\hline
STMC~\cite{zhou2020spatial}* & 2.1 \\
\hline
Baseline & 3.5   \\
Baseline+VAC & \textbf{1.6}  \\\shline
\end{tabular}
\label{tab: csl}
\vspace{-10 pt}
\end{table}

\subsection{Comparison with the State-of-the-art.}
We present the comparison results with several state-of-the-art approaches in Table~\ref{tab: phoenix} and Table~\ref{tab: csl}. From Table~\ref{tab: phoenix} we can see that the proposed method with gloss-wise temporal layer and VAC achieves competitive results with previous iteration-based methods. We can also illustrate the success of STMC~\cite{zhou2020spatial} and CMA~\cite{pu2020boosting} from the overfitting perspective: the former enforces the feature extractor to extract visual information from extra supervision and the latter weakens the contextual information with the data augmentation.

To examine the generalization of the proposed method, we also evaluate it on the CSL dataset. As no official split is given, the performance comparison among methods in Table~\ref{tab: csl} has limited practical value. The proposed method shows improvement than baseline and achieves better performance than recent work~\cite{cheng2020fully} under the same setting.

\subsection{Discussion}
\label{sec:dis}
We can roughly divide recent methods into two categories from the overfitting perspective: enhancing the feature extractor~\cite{cheng2020fully,cui2019deep,pu2019iterative,koller2019weakly,yang2019sf,zhou2020spatial} and weakening the alignment module~\cite{cheng2020fully,koller2017re,niustochastic}. The proposed VAC is an attempt to make better use of visual information, which provides a new perspective to solve this problem. How to better use visual features with a more powerful temporal model, which will be easier to overfit but can further improve WAR, is a challenging problem.


\section{Conclusion}
\label{sec:con}
Overfitting is one of the major problems in CTC-based sign language recognition, which leads to insufficient training of the feature extractor. In this study, we propose the visual alignment constraint to make CSLR networks end-to-end trainable by enforcing the feature extractor to make predictions with more alignment supervision. Two metrics are proposed to measure the inconsistent predictions of the feature extractor and the alignment module. Experimental results show that the proposed VAC narrows the gap between predictions of the auxiliary and the primary classifiers. The proposed metrics and relevant experiments provide a new perspective on the relationship between visual and alignment modules, and we hope they can inspire future studies on CSLR and other sequence classification tasks.

Our source codes and trained models are available at \\ \url{https://vipl.ict.ac.cn/resources/codes} or \url{https://github.com/ycmin95/VAC_CSLR}. 

\medskip
\noindent\textbf{Acknowledgement.} This study was partially supported by the Natural Science Foundation of China under contract No.  61976219.

{\small
\bibliographystyle{ieee_fullname}
\bibliography{egbib}
}

\appendix
\cleardoublepage

{\section*{\Large Appendix}}

This appendix provides details that are not shown in the main paper. We first present the training process of the proposed VAC (\S~\ref{sec: train}) and ablations on dataset size (\S~\ref{sec: data}), temporature (\S~\ref{sec: tem}), loss weight (\S~\ref{sec: wei}) and augmentation (\S~\ref{sec: aug}). Then we present the details of the temporal convolution designs (\S~\ref{sec: net}), the proposed metrics (\S~\ref{sec:  metric}) and the performance gap (\S~\ref{sec:gap}). Finally, we visualize the spatial activations (\S~\ref{sec: rec}), magnitudes (\S~\ref{sec: mag}) and more qualitative results (\S~\ref{sec: qua}).



\section{Additional Results}
\subsection{Training process of VAC}
\label{sec: train}
\begin{figure}[h]
\centering
\includegraphics[width=0.8\linewidth]{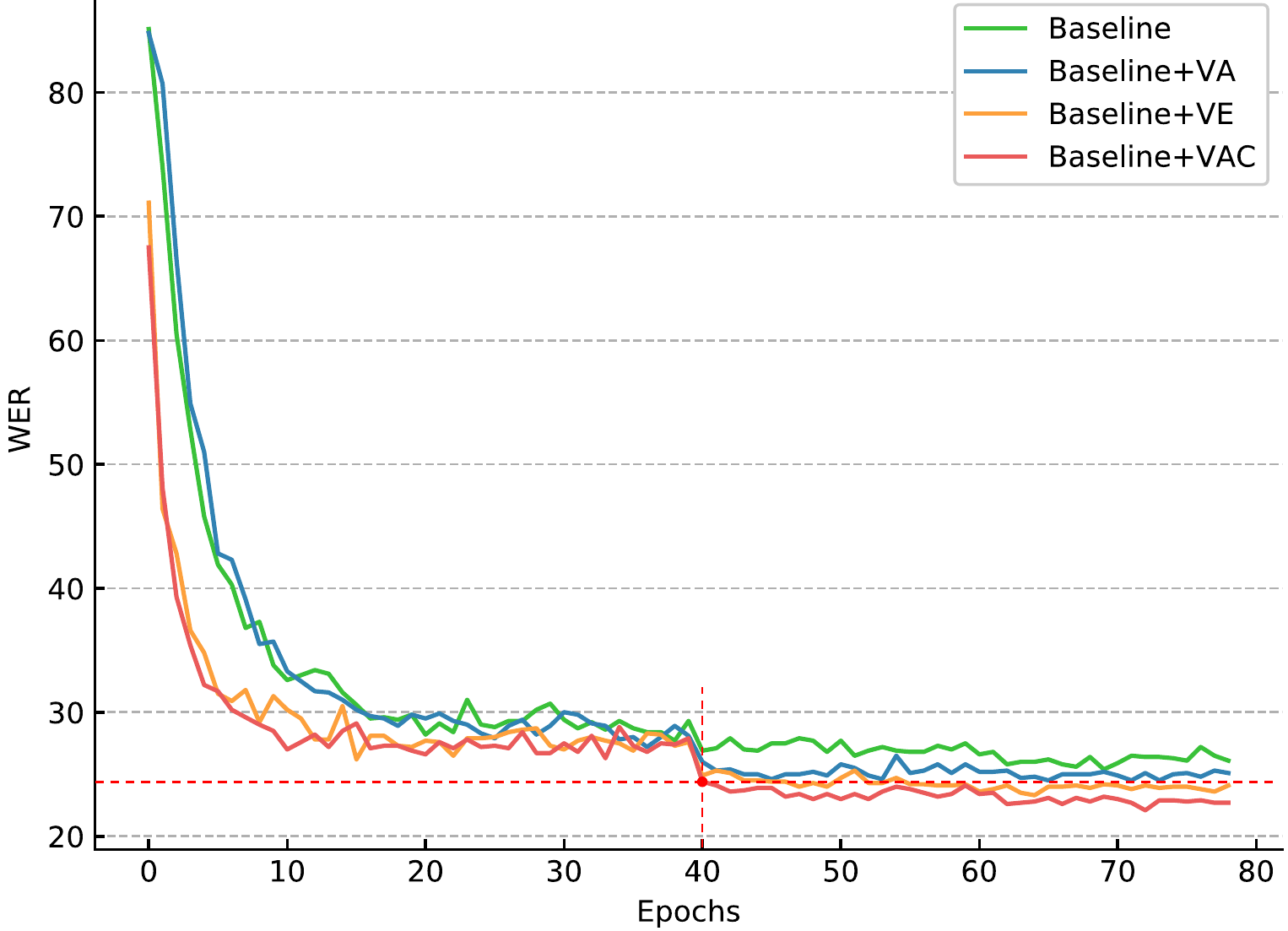}
\caption{ Learning curves of WER(\%) on PHOENIX14 with different settings. The learning rate is decayed at 40 and 60 epochs.} 
\label{fig:training}
\end{figure}

We compare the curves with different constraints in Fig.~\ref{fig:training}. Adopting VAC can significantly accelerate the training process, which achieves better performance than baseline after the first learning rate decay. The $\mathcal{L}_{VE}$ can immediately accelerate the training process at the beginning and the $\mathcal{L}_{VA}$ takes effect when the alignment model begins to converge, which happens after the first learning rate decay.

\subsection{Ablation on Dataset Size}
\label{sec: data}
We visualize the recognition results with different sizes of training data in Fig.~\ref{fig:size} below. It can be seen that VAC can steadily improve performance as the training data size increases, while the visual extractor of the baseline ($\text{WER}_a$) shows a saturation trend, which implies the available training data is \textbf{NOT} sufficient for the visual extractor.

\begin{figure}[h]
\vspace{-12pt}
\centering
\includegraphics[width=0.72\linewidth]{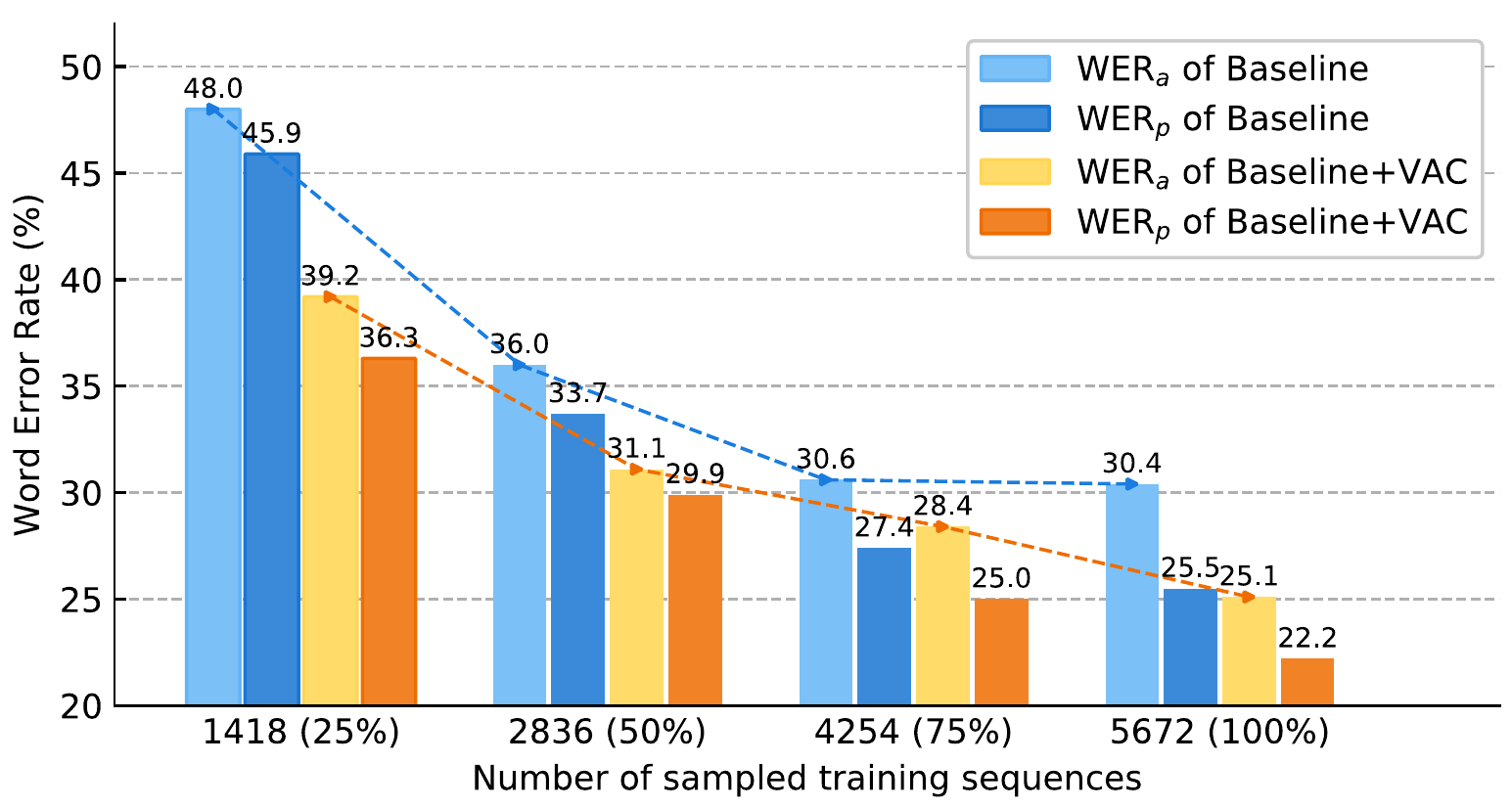}
\caption{Results on PHOENIX14 with sampled training set.} 
\label{fig:size}
\vspace{-10pt}
\end{figure}

\subsection{Ablation on Temporature $\tau$}
\label{sec: tem}
To determine the temperature $\tau$ in Equ. 5 of the main paper, we evaluate its effect in Table~\ref{tab: temp}. Low temperatures leads to spike responces and high temperatures will produce noisy supervision. According to ablation results, $\tau$=8 is a proper choice.

\begin{table}[h]
\centering
\setlength{\tabcolsep}{11pt}
\caption{Ablation results (WER, \%) of temporature $\tau$.}
\begin{tabular}{cccccc}  \shline
$\tau$ & 1 & 4 & 8 & 12 & 16 \\  \hline
Dev & 22.1 & 22.0 & \textbf{21.2}  & 21.7 & 21.6 \\
Test & 22.8 & 22.9 & \textbf{22.3} & 22.9 & 22.7 \\  \shline
\end{tabular}
\label{tab: temp}
\end{table}

\subsection{Ablation on Loss Weight $\alpha$}
\label{sec: wei}

\begin{table}[h]
\centering
\setlength{\tabcolsep}{8pt}
\caption{Ablation results (WER, \%) of loss weight $\alpha$.}
\begin{tabular}{ccccccc} \shline
$\alpha$ & 10 & 15 & 20 & 25 & 30 & 35 \\ \hline
Dev & 22.1 & 21.9 & 21.5 & \textbf{21.2} & 21.5 & 22.0 \\ 
Test & 23.0 & 22.4 & \textbf{22.1} & 22.3 & 22.6 & 23.2 \\ \shline
\end{tabular}
\label{tab: loss}
\end{table}

Another hyperparameter need to be carefully tuned is the loss weight $\alpha$ in Equ. 6 of the main paper. We conduct ablation study on it and present results in Table~\ref{tab: loss}. As weight of distillation increases, the performance first increases and then decreases after certain value. The optimal weight for distillation loss is 25.

\subsection{Ablation on Data Augmentation}
\label{sec: aug}
As mentioned in Sect. 5.1, we adopt three kinds of data augmentation strategies (random crop, horizontal flip and random temporal scaling) during training, which is the same as previous work~\cite{zhou2020spatial}. In Table~\ref{tab: aug}, we evaluate the effect of data augmentation. We can observe that adopting data augmentation can significantly improve the performance, especially with random crop. We assume that the network has a tendency to use shortcuts, such as the absolute position of hands in video, and adopting random crop can enforce the network to learn more high-level features and mitigate these shortcuts. It is interesting to see that the horizontal flip can improve the results although all signers in PHOENIX14 use their right hand as the dominant hand when signing, which brings about 0.6\% performance gain.

\begin{table}[h]
\setlength{\tabcolsep}{9pt}
\caption{Ablation results (WER, \%) of augmentation.}
\begin{tabular}{ccccc}
\shline
Crop & Flip & Temporal Scaling & Dev & Test \\\hline
 &  &  & 28.1 & 28.4 \\
\checkmark &  &  & \textbf{23.8} & \textbf{24.6} \\
 & \checkmark &  & 26.1 & 26.4 \\
 &  & \checkmark & 27.4 & 27.3 \\\hline
\checkmark & \checkmark &  & 23.2 & 23.8 \\
\checkmark & \checkmark & \checkmark & \textbf{22.1} & \textbf{23.0} \\
\shline
\end{tabular}
\label{tab: aug}
\end{table}

\begin{table*}[h]
\centering
\setlength{\tabcolsep}{12pt}
\caption{More details about the temporal layer design. Conv1x$\alpha$ (1x$\alpha$ Convolution-BN-ReLU) and Max-pooling 1x$\beta$ are used to extract different levels of features.}
\begin{tabular}{cccccl}
\shline
 & \multicolumn{4}{c}{Layer} & Output Size \\ \hline
Backbone & \multicolumn{4}{c}{ResNet18} & $(B, C, 1, T)$ \\ \hline
\multirow{5}{*}{Temporal Layer} & \multicolumn{2}{c|}{Framewise} & \multicolumn{1}{c|}{Subgloss wise} & Gloss wise & \multirow{5}{*}{$(B, C', 1, T')$} \\ \cline{2-5}
 & \multicolumn{1}{p{1.2cm}<{\centering}|}{Conv1x1} & \multicolumn{1}{p{1.2cm}<{\centering}|}{Conv1x3} & \multicolumn{1}{c|}{Conv1x5} & Conv1x5 &  \\ 
 & \multicolumn{1}{p{1.2cm}<{\centering}|}{} & \multicolumn{1}{p{1.2cm}<{\centering}|}{} & \multicolumn{1}{c|}{Max-pooling 1x2} & Max-pooling 1x2 &  \\ 
 & \multicolumn{1}{p{1.2cm}<{\centering}|}{} & \multicolumn{1}{p{1.2cm}<{\centering}|}{} & \multicolumn{1}{c|}{} & Conv1x5 &  \\ 
 & \multicolumn{1}{p{1.2cm}<{\centering}|}{} & \multicolumn{1}{p{1.2cm}<{\centering}|}{} & \multicolumn{1}{c|}{} & Max-pooling 1x2 &  \\ \hline
\multirow{2}{*}{Alignment model} & \multicolumn{4}{c}{BiLSTM$(C', 512, 2)$} & \multirow{2}{*}{$(B, T', N)$} \\ \cline{2-5}
 & \multicolumn{4}{c}{Linear$(1024, N)$} &  \\ \shline
\end{tabular}
\label{lab: net}
\end{table*}

\section{Additional Implementation Details}

\subsection{Details on Temporal Layer Designs}
\label{sec: net}


\begin{table*}[t]
\centering
\caption{Train/Dev/Test performance comparison (\%) with different evaluate metrics on PHOENIX14. $\text{WER}_p^*$ and $\text{WER}_a^*$ correspond to the $\text{WER}^*$ results of primary classifier and auxiliary classifier, respectively, and $\Delta \text{WER}^* = \text{WER}_a^* - \text{WER}_p^* = \text{WAR}  - \text{WDR}$.}
\begin{tabular}{lccccc}
\shline
 & $\text{WER}_a^*$ & $\text{WER}_p^*$ & WAR & WDR & $\Delta\text{WER}^*$  \\\hline
Baseline & 12.9 / 30.4 / 29.4 &  2.5 / 25.5 / 26.9 & 11.5 / 11.3 / 10.0 & 1.2 / 6.5 / 7.4 & 10.4 / 4.9 / 2.5 \\
Baseline + iteration & \ \ 7.9 / 27.5 / 27.0  & 1.9 / 25.1 / 26.3 & \ \ 7.0 / 9.8 / 8.8 & 0.9 / 7.3 / 8.2 &\ \ 6.0 / 2.4 / 0.7 \\\hline
Baseline + VE & \ \ 3.8 / 26.2 / 26.3 & 2.5 / 23.4 / 24.0 & \ \ 2.1 / 6.2 / 5.8 & 0.8 / 3.4 / 3.4  &\ \  1.3 / 2.8 / 2.3  \\
Baseline + VA & 13.4 / 26.8 / 26.9 & 2.0 / 24.7 / 25.2 & 12.0 / 7.7 / 7.8 & 0.6 / 5.7 / 6.2 & 11.4 / 2.1 / 1.7 \\ 
Baseline + VAC & \ \ 3.9 / 25.1 / 25.2 & 1.9 / 22.2 / 23.0 & \ \ 2.3 / 5.5 / 5.0 & 0.4 / 2.6 / 2.8 &\ \ 2.0 / 2.9 / 2.2 \\\shline
\end{tabular}
\label{tab: comp}
\vspace{-10pt}
\end{table*}

As mentioned in Sect. 5.2, we evaluate three kinds of basic temporal convolution layers and present the details in Table~\ref{lab: net}. The output dimension $C$ of the ResNet18~\cite{he2016deep}  is 512, and the output dimension $C'$ of the temporal layer is 1024. Conv1x$\alpha$ (1x$\alpha$ Convolution-BN-ReLU) and Max-pooling 1x$\beta$ are used to extract different levels of features. The lengths $T'$ of output sequences of (Frame-wise Raw, Frame-wise Conv1x3, Subgloss-wise, Gloss-wise) are $(T, T-2, T/2-2, T/4-3)$. The alignment model contains a two-layer BiLSTM (512 hidden states for each direction) and a fully-connected layer with $N$ output units is adopted to make the final prediction.

\subsection{Details on Proposed Metrics}
\begin{figure}[ht]
\centering
\subfigure[Alignment process of three sentences.]{
\begin{minipage}[t]{1\linewidth}
\includegraphics[width=1.0\linewidth,page=1]{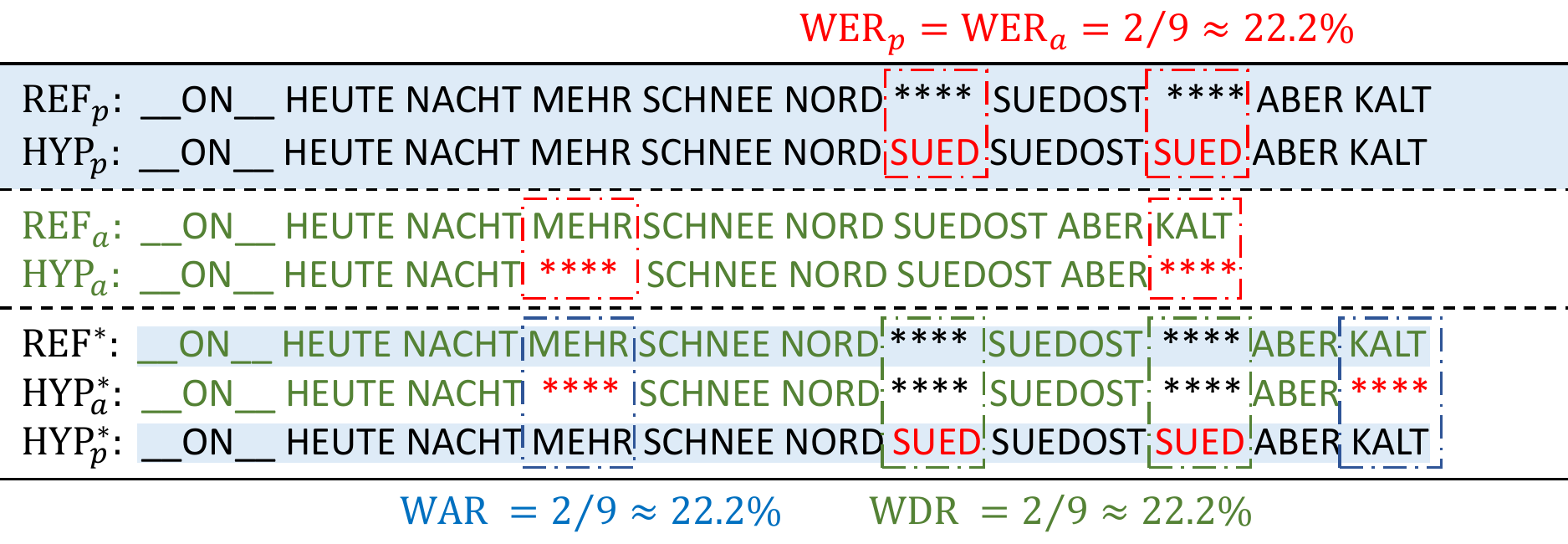}
\label{fig:example}
\end{minipage}%
}%
\vspace{-8pt}
\subfigure[An example of performance deterioration of the primary classifier.]{
\begin{minipage}[t]{1\linewidth}
\includegraphics[width=1.0\linewidth,page=2]{imgs/supp_figure9_compare.pdf}
\label{fig:pri}
\end{minipage}
\vspace{-130pt}
}%
\vspace{-8pt}
\subfigure[An example of performance amelioration of the primary classifier.]{
\begin{minipage}[t]{1\linewidth}
\includegraphics[width=1.0\linewidth,page=3]{imgs/supp_figure9_compare.pdf}
\vspace{-25pt}
\label{fig:hhh}
\end{minipage}
}%
\caption{ Alignment results of the proposed alignment method. We highlight {\color{red}{wrong recognition glosses}} and the alignment results of \textcolor[rgb]{0.3254,0.5098,0.2078}{the auxiliary classifier}, \colorbox[rgb]{0.8705,0.9216,0.9686}{\color{black}the primary classifier}.}

\label{fig:examples}
\end{figure}

\label{sec: metric}
In Sect. 4.2, we propose two metrics, Word Deterioration Rate (WDR) and Word Amelioration Rate (WAR), to evaluate the performance of the recognition results. To calculate WDR and WAR, we need to align the reference sentence and the recognized sentences from the auxiliary classifier and the primary classifier first. As shown in Fig.~\ref{fig:example}, we first align the reference and the recognized sentences and refer the alignment results as to ($\text{REF}_p$, $\text{HYP}_p$) and ($\text{REF}_a$, $\text{HYP}_a$) for the primary classifier and the auxiliary classifier, respectively. Then we align $\text{REF}_a$ and $\text{REF}_p$ to obtain the aligned reference $\text{REF}^*$. The final alignment results ($\text{REF}^*$, $\text{HYP}^*_a$, $\text{HYP}^*_p$) are presented in the last row of Fig.~\ref{fig:example} by aligning ($\text{REF}^*$, $\text{HYP}_a$) and ($\text{REF}^*$, $\text{HYP}_p$), respectively.

With the help of alignment results, we can compare the performance of the two classifiers. As shown in Fig.~\ref{fig:example}, both of the auxiliary and the primary classifiers have the same WER 22.22\% ($\text{HYP}_p$ has two insertion errors, and $\text{REF}_a$ has two deletion errors). The primary classifier corrects the misrecognized results of the auxiliary classifier but makes new mistakes, which can not be measured by WER. WDR measures the ratio that is correctly recognized by the auxiliary classifier but misrecognized by the primary classifier (two `SUED' in $\text{HYP}^*_p$), and WAR does in the opposite direction (`MEHR' and `KALT' in $\text{HYP}^*_p$). Based on the proposed metrics, we can calculate that both WAR and WDR are 22.22\% and better understand the recognition results: the introduction of the alignment model brings 22.22\% gains and extra 22.22\% errors, so the total WER remains unchanged.

Due to the alignment process and different weights of operations, the proposed three-sentence alignment strategy leads to a little performance degradation than the general WER, as discussed in Sect. 5.2. Fig.~\ref{fig:pri} and Fig.~\ref{fig:hhh} show some examples. Aligning $\text{REF}_a$ and $\text{REF}_p$ changes the alignment results, which often breaks substitution errors to more deletion and insertion errors. However, only a small ratio of sequences has such a problem, and we believe this problem is acceptable for results analysis.

\subsection{Details on the Performance Gap}
\label{sec:gap}
Figure 6 in Sect. 5.2 visualizes the performance gap with different settings, and we present the detailed results in Table ~\ref{tab: comp}. The conclusions in Sect. 5.2 are consistent on both dev and test sets.

\section{Qualitative Results}

\subsection{Visualization of Spatial Activations}
\label{sec: rec}
\begin{figure}[!ht]
\centering
\includegraphics[width=0.9\linewidth]{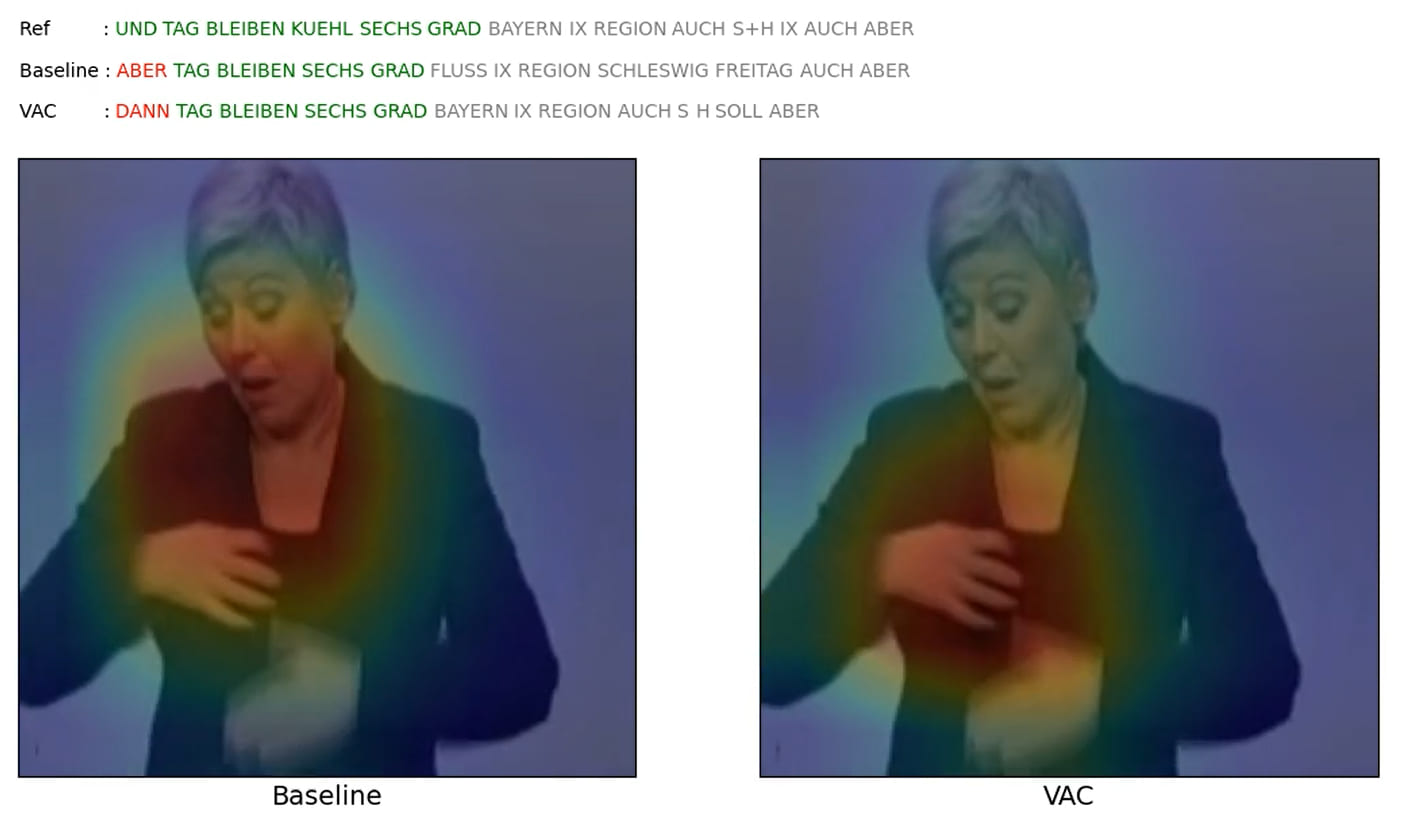}
\label{fig:aux}
\caption{ Interface of the recognition result animation. We highlight the \color{red}{wrong recognition glosses}.} 
\label{fig:demo}
\end{figure}

We visualize some recognition results in the animation folder. As shown in Fig.~\ref{fig:demo}, the reference and the predictions of baseline and Visual Alignment Constraint (VAC) are presented above the videos. The bottom videos visualize the activation changes during the signing. The activation maps are obtained by calculating the $l_2$ norm of the 7x7 ResNet18 feature maps. From the animation, we can observe that the baseline mainly focuses on the central area of frames, and the proposed method can dynamically focus on hands and facial expressions, which extracts more discriminative visual features.

\subsection{Visualizing of Magnitudes}
\label{sec: mag}
\begin{figure*}[t]
\subfigure[Baseline on training set]{
\begin{minipage}[t]{1\linewidth}
\begin{center}
\includegraphics[width=0.9\linewidth,page=1]{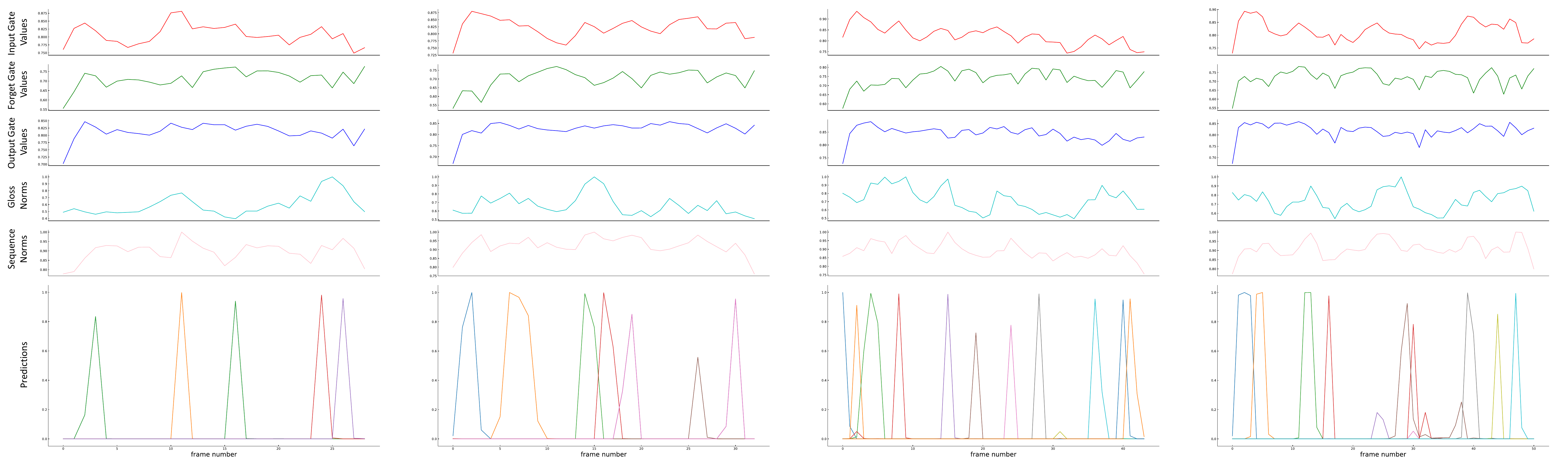}
\end{center}
\end{minipage}%
}%

\subfigure[Baseline on dev set]{
\begin{minipage}[t]{1\linewidth}
\begin{center}
\includegraphics[width=0.9\linewidth,page=3]{imgs/supp_figure11_norm_vis.pdf}
\end{center}
\end{minipage}%
}%

\subfigure[Baseline+VAC on training set]{
\begin{minipage}[t]{1\linewidth}
\begin{center}
\includegraphics[width=0.9\linewidth,page=2]{imgs/supp_figure11_norm_vis.pdf}
\end{center}
\end{minipage}%
}%

\subfigure[Baseline+VAC on dev set]{
\begin{minipage}[t]{1\linewidth}
\begin{center}
\includegraphics[width=0.9\linewidth,page=4]{imgs/supp_figure11_norm_vis.pdf}
\end{center}
\end{minipage}%
}%
\caption{Visualization of the gate values, the $l_2$ norm of features and the final prediction on PHOENIX14.}
\label{fig:vis}
\vspace{-10pt}
\end{figure*}

In Sect. 3.3, we propose a magnitude hypothesis that the $l_2$ norms of features reflect the importance of frames. Besides, experimental results in Sect. 5.2 verify that the proposed VAC is more compatible with the spiky activations. Fig.~\ref{fig:vis} presents the gate values, the $l_2$ norms of features, and the final predictions on dev and training sets. The baseline shows different behavior on training and dev sets: the norms of gloss and sequence features have consistent tendencies on the training set but the correlations become weakened on the dev set. Baseline+VAC shows consistent behavior on both sets, which indicates the effectiveness of the proposed VAC.

\subsection{More Qualitative Recognition Results}
\label{sec: qua}
\begin{figure*}[t]
\subfigure[The primary classifier provides better results than the auxiliary.]{
\begin{minipage}[t]{1\linewidth}
\begin{center}
\includegraphics[width=0.9\linewidth,page=1]{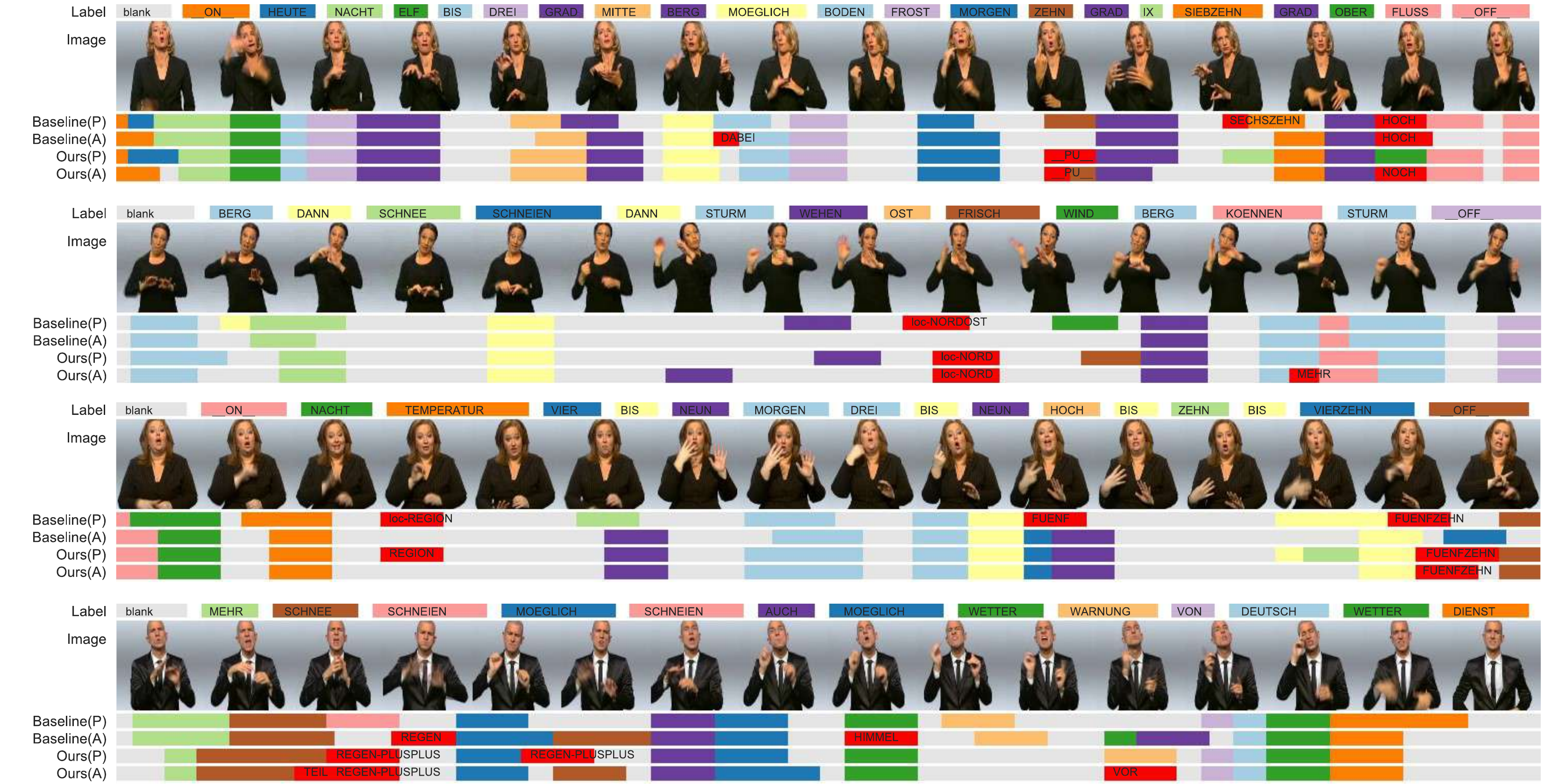}
\end{center}
\label{fig:visual_pri}
\end{minipage}%
}%

\subfigure[The auxiliary classifier provides better results than the primary.]{
\begin{minipage}[t]{1\linewidth}
\begin{center}
\includegraphics[width=0.9\linewidth,page=2]{imgs/supp_figure12_visual.pdf}
\end{center}
\vspace{-3pt}
\label{fig:visual_aux}
\end{minipage}
}%

\caption{Qualitative comparison among different network settings with examples from Dev set on PHOENIX14. Wrong recognition results (except deletion operations) are marked in red. The primary classifier and auxiliary classifier outputs are marked as (P) and (A).}
\label{fig:visualize}
\vspace{-10pt}
\end{figure*}

We visualize more sequences in Fig.~\ref{fig:visualize}, and we can notice that the prediction results of two classifiers are not always consistent. As shown in Fig.~\ref{fig:visual_pri}, the primary classifier can provide better results by incorporating more context information. However, the primary classifier may neglect visual information or predict wrong glosses, which gives worse results in some cases, as shown in Fig.~\ref{fig:visual_aux}. The proposed VAC attempts to make better use of visual and context information.

\end{document}